\documentclass[nohyperref]{article}

\usepackage{microtype}
\usepackage{graphicx}
\usepackage{subfigure}
\usepackage{booktabs} 

\usepackage{hyperref}

\usepackage[accepted]{icml2022}

\usepackage{amsmath}
\usepackage{amssymb}
\usepackage{mathtools}
\usepackage{amsthm}

\usepackage[capitalize,noabbrev]{cleveref}

\theoremstyle{plain}

\theoremstyle{definition}

\theoremstyle{remark}

\usepackage[textsize=tiny]{todonotes}

\icmltitlerunning{Generating Kernels for Quantized LLM Inference}

\usepackage{amssymb}
\usepackage{amsopn}
\usepackage{algorithm}
\usepackage{tikz}
\usepackage{caption}

\newcommand{\mypar}[1]{\textbf{#1}.}

\newcommand{\bits}{\text{b}}
\newcommand{\R}{\mathbb{R}}
\newcommand{\N}{\mathbb{N}}
\newcommand{\round}{\operatorname{rnd}}
\newcommand{\zeros}{\text{z}}
\newcommand{\scale}{\text{s}}
\newcommand{\avx}[1]{\textbf{\texttt{#1}}}

\newcommand{\ra}[1]{\renewcommand{\arraystretch}{#1}}

\begin{document}

\twocolumn[
\icmltitle{QIGen: Generating Efficient Kernels for Quantized Inference \\ on Large Language Models}



\icmlsetsymbol{equal}{*}

\begin{icmlauthorlist}
\icmlauthor{Tommaso Pegolotti}{eth}
\icmlauthor{Elias Frantar}{ista}
\icmlauthor{Dan Alistarh}{istanm}
\icmlauthor{Markus P\"uschel}{eth}
\end{icmlauthorlist}

\icmlaffiliation{eth}{ETH Zurich}
\icmlaffiliation{ista}{IST Austria}
\icmlaffiliation{istanm}{IST Austria \& Neural Magic}

\icmlcorrespondingauthor{Tommaso Pegolotti}{tommaso.pegolotti@inf.ethz.ch}
\icmlcorrespondingauthor{Dan Alistarh}{dan.alistarh@ist.ac.at}


\vskip 0.3in
]



\printAffiliationsAndNotice{} 

\begin{abstract}
We present ongoing work on a new automatic code generation approach  for supporting quantized generative inference on LLMs such as LLaMA or OPT on off-the-shelf CPUs. Our approach is informed by the target architecture and a performance model, including both hardware characteristics and method-specific accuracy constraints. Results on CPU-based inference for LLaMA models show that our approach can lead to high performance and high accuracy, comparing favorably to the best existing open-source solution. A preliminary implementation is available at \url{https://github.com/IST-DASLab/QIGen}.
\end{abstract}

\vspace{-1.5em}
\section{Introduction}

The impressive performance of generative large language models (LLMs)~\cite{black2022gpt, zhang2022opt, touvron2023llama} has led to significant interest in executing them on user devices with limited computational power. 
Interestingly, the computational envelope of generative workloads renders this plausible: when executing the popular one-token-at-a-time generation based on a given cached context, the operational cost of obtaining single outputs is relatively small, as the computation can be mapped onto relatively inexpensive matrix-vector products, as opposed to the massive matrix-matrix multiplications that are otherwise common in deep learning. 

Thus, the key challenge in personalized generative LLM inference becomes \emph{memory}: well-performing models have extremely large parameter counts, which often exceed the memory capacity of consumer devices, and induce high memory transfer costs at runtime, overwhelming bandwidth. 
To address this issue, a series of \emph{quantization-based methods} specialized to LLMs have been recently proposed~\cite{dettmers2022llm, dettmers2022case, frantar2022gptq, park2022nuqmm, xiao2022smoothquant, yao2022zeroquant}, which work by reducing the bitwidth of data types used for storing weights, activations, or both, with the goal of minimizing the impact on accuracy. 

Focusing specifically on generative inference, where the size of the weights is the main bottleneck, the currently best-performing method is GPTQ~\cite{frantar2022gptq}, which achieves near-lossless quantization to 4-bit weights, and can even accurately support 2 and 3-bit weights by reducing the granularity to smaller weight groups, e.g., by jointly quantizing blocks of 64 weights using a shared scale and zero-point. Similar grouping techniques can also allow simpler round-to-nearest (RTN) quantization to preserve accuracy 
using 4-bit weights~\cite{dettmers2022case}. 

Given this algorithmic progress, a remaining key challenge is the efficient system support for these compressed numerical formats, to execute LLMs on user devices accurately and fast.  
Existing academic proposals such as LLM.int8()~\cite{dettmers2022llm}, GPTQ~\cite{frantar2022gptq}, and  SmoothQuant~\cite{xiao2022smoothquant}, and open-sourced solutions such as llama.cpp~\cite{llamacpp}, manually develop custom kernels for their specific target hardware, such as GPUs or CPUs. 
Unfortunately, this approach can be extremely time-intensive and error-prone, and requires potentially re-writing kernels from scratch to support new quantization formats and target hardware. 

In this paper, we present ongoing work on a new automatic code generation approach, called QIGen, for obtaining efficient and general kernels for generative LLM inference of varying bitwidth. At a high level, our approach provides customized efficient implementations of the low-level matrix operations required to support multiplication operations on quantized variants of LLM weight matrices. Our approach is based on a performance model which is informed both by hardware characteristics, e.g., cache size, and by accuracy constraints pertaining to the quantization methods, e.g., the use of weight grouping. We present results generating efficient low-bitwidth kernels for general purpose CPUs supporting the popular AVX2 intrinsics, which we interface with Pytorch~\cite{paszke2019pytorch}, and  showcase on the accurate LLaMA family~\cite{touvron2023llama}.

\section{Background}

\mypar{Quantization}
Quantization is an efficient compression technique for reducing memory utilization by representing data using a limited number of values, typically integer levels. We define a \emph{quantization function} as a map $Q$ from real numbers to integers. Formally,
\begin{displaymath}
Q : (\R, \N) \mapsto [0, 2^\bits),
\end{displaymath} where $\bits$ is the number of bits we want to use to represent the new value. Given a vector $x \in \R^n$, we define $Q(x,\bits)$ as
\begin{displaymath}
Q(x,\bits) = \round \Bigg(\frac{x - \min(x)}{\max(x) - \min(x)} (2^\bits -1) \Bigg),
\end{displaymath}
where $\max(x)$ and $\min(x)$ is the maximum and minimum value in $x$, respectively, and $\round$ rounds to nearest. This equation can be rewritten as $x_q = \round((x - \zeros)\scale(x))$, with $\zeros = \min(x)$ and $\scale(x) = (2^\bits -1) / (\max(x) - \min(x))$. Similarly, the corresponding dequantization function is
\begin{displaymath}
D(x_q) = \scale (x_q  \zeros).
\end{displaymath}

As example, consider the dot-product between $y \in \R^n$ and $x_q \in [0, 2^\bits)$. The resulting value is given by
\begin{equation}\label{eq:core}
\langle y,  \scale (x_q - \zeros) \rangle = \scale (\langle y, x_q \rangle - \zeros  \langle y, 1 \rangle).
\end{equation}

To improve accuracy, we can increase the quantization granularity and store more $\scale$ and $\zeros$ values for each vector. We denote the resulting partition by groups. To compute the dot-product using these additional values, we only need to modify~\ref{eq:core}. In particular, we rewrite it as
\begin{equation}\label{eq:core_group}
\small\sum_i^P \langle y_i,  \scale_i (x_{iq} - \zeros_i) \rangle = \small\sum_i^P\scale_i \langle y_i, x_{iq} \rangle - \small\sum_i^P \zeros_i  \langle y_i, 1 \rangle,
\end{equation}
where $P$ is the number of groups.


\mypar{LLM quantization}
There has been significant focus on accurate post-training quantization (PTQ) methods~\cite{nagel2019data} that scale and are accurate for LLMs. 
Early work~\citep{yao2022zeroquant, dettmers2022llm, park2022nuqmm} 
used direct rounding to the nearest quantization level (RTN), 
reducing group size to obtain higher accuracy at the cost of more space. LLM.int8()~\citep{dettmers2022llm} quantized activations as well, isolating ``outlier features'' for which higher bit-width is used. 
These approaches induce quantization errors of 5--10\% in perplexity increase for OPT~\cite{zhang2022opt} or LLaMA~\cite{touvron2023llama} models, relative to the uncompressed baseline. 
GPTQ~\cite{frantar2022gptq} proposed a higher-accuracy approach (e.g., 3--5\% perplexity increase at 4-bit), via an approximate solver minimizing the layer-wise squared error between the quantized and original layers.  
\citet{dettmers2022case} provided an in-depth overview of the accuracy-compression trade-offs underlying these methods, 
establishing that 4-bit quantization is an optimal point for round-to-nearest-based methods,  
whereas higher compression can be achieved via data-aware methods such as GPTQ. 

All the above methods focused on GPU inference as their main target scenario. 
By contrast, there has been much less focus on \emph{CPU-based inference}; 
for this, the open-source LLaMA.cpp/GGML project~\cite{llamacpp} can provide reasonable generative performance on end devices such as Intel/AMD/ARM CPUs, showing that running models locally in such setups is feasible. 

\mypar{Notation}
We simplify the names of AVX SIMD vector instructions for readability in our exposition. The semantics of the instructions can be seen in~\cref{tbl:instructions}.

\begin{table}[t]
\centering
\ra{1.3}
\footnotesize
\begin{tabular}{@{}ll@{}}\toprule
\avx{load}(\texttt{address}) & load 
from memory address \\
\avx{store}(\texttt{address,a}) & stores \texttt{a} at memory address \\
\avx{broadcast}(\texttt{a}) & fills a register with \texttt{a}\\
\avx{fmadd}(\texttt{a,b,c}) & returns $a \cdot b + c$\\
\avx{reduce\_add}(\texttt{a}) & returns the sum of the elements of \texttt{a}\\
\avx{srli}(\texttt{a,i}) & returns $\texttt{a} >> i$\\
\avx{and}(\texttt{a,m}) & returns $\texttt{a} \text{ bitwise and } m$\\
\avx{cvt\_int\_float}(\texttt{a}) & converts \texttt{a} to float\\
\bottomrule
\end{tabular}
    \caption{List of vector instructions used in the implementation and their semantics.}\label{tbl:instructions}
\end{table}

\section{Code Generation}

We consider CPU-based generative inference as the motivating setup for our work, although our techniques are general, and should be extensible to other settings as well. 

\mypar{Data format}
Our implementation of linear layers uses quantized weights obtained via GPTQ, in which the weights are represented using 32-bit integers to store multiple consecutive values. For instance, with 4-bit quantization, a single ``unit'' can represent 8 values, and with 2-bit quantization, it can represent 16 values. Some bit granularities may require specialized implementations. For example, if we use 3-bit, we store 32 values in three consecutive integers. 

\begin{figure}
    \centering
    \includegraphics[scale=0.8]{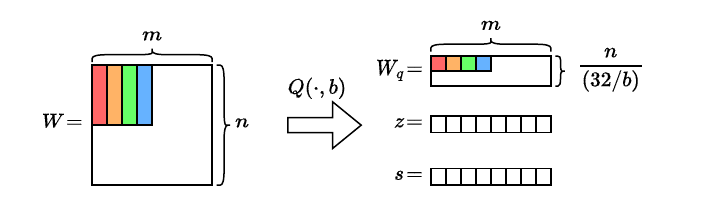}
    \caption{Reduction in matrix size due to quantization. Given a weight matrix $W$ of size $n\times m$, we obtain a compressed matrix $W_q$ of size $n/(32/\bits) \times m$ and two one dimensional vectors $\zeros$ and $\scale$ of size $m$. We quantize the elements column-wise as shown by the colors. For example, $8$ values in the red rectangle become $1$ value in the red square.}
    \label{fig:quant}
\end{figure}

We quantize the weight matrices in a column-wise manner. For 4-bit quantization, each entry $(i,j)$ in the quantized matrix contains the values at indices $(8i:8i+7, j)$ of the uncompressed matrix. For each column $j$, we also store one scale $\scale_j$ and one $\zeros_j$ value. Additionally, we can also quantize the $\zeros_j$ value. The changes in the matrices after quantization are illustrated in~\cref{fig:quant}.

The total memory used to store a weight matrix of size $N \times M$ is thus $\bits N M + 32 M + \bits M$ instead of $32NM$. The $32M$ is for the $\scale_j$ stored in single-precision floating point. 
We obtain a significant reduction in memory usage by a factor of $\approx 32 / \bits$.

\mypar{Computation}
LLMs typically comprise a series of linear layers, where the input is a vector. As a result, matrix-vector  multiplications (GEMV) form the core of our computations. In Algorithm~\ref{alg:qgemv}, we present a straightforward implementation of a 4-bit qGEMV (quantized general matrix-vector multiply) based on the factorization in~\eqref{eq:core}. The approach involves computing the dot-products using integer representations and scaling and transposing the result once after the final reduction. To extract eight rows from the quantized weight matrix, considering 4-bit quantization, we utilize the auxiliary function \avx{unpack} as presented in Algorithm~\ref{alg:deq}.



\begin{algorithm}
\footnotesize
\caption{4-bit qGEMV routine.}\label{alg:qgemv}
\begin{algorithmic}[1]
\STATE {\bfseries In: } $W \in \N^{\frac{n}{\bits}\times m}$, $s \in \R^m$, $z \in \R^m$, $x \in \R^n$, $\hat x \in \R$
\STATE {\bfseries Out: } $y = xW$
\FOR{$j = 0:m$}
\STATE $acc \gets \avx{broadcast}(0)$
\FOR{$i = 0:8:n$}
\STATE $w \gets \! \avx{load}(\&W[i/8][j])$
\STATE $l_{0:8} \gets \avx{unpack}(w)$
\FOR{$ii = 0:8$}
\STATE $x_{ii} \gets \avx{broadcast}(x[i+ii])$
\STATE $acc \gets \avx{fmadd}(x_{ii},l_{ii},acc)$
\ENDFOR
\ENDFOR
\STATE $w[j] \gets s[j] * (\avx{reduce\_add}(acc) - z[j] * \hat x)$
\ENDFOR
\end{algorithmic}
\end{algorithm}

\begin{algorithm}
\caption{AVX2 4-bit Unpack Routine.}\label{alg:deq}
\begin{algorithmic}[1]
\small
\STATE {\bfseries In: } $v$
\STATE {\bfseries Out: } $l_0$ to $l_7$
\STATE $\text{ mask } \gets \avx{broadcast}(15)$
\FOR{$i = 0:8$}
\STATE $s_i \gets \avx{srli}(v, i*4)$
\STATE $a_i \gets \avx{and}(s_i, \text{mask})$
\STATE $l_i \gets \avx{cvt\_int\_float}(a_i)$
\ENDFOR
\end{algorithmic}
\end{algorithm}

\mypar{Mini-GEMV}
To improve performance, we exploit that weight matrices in a neural network are set at compile time. By considering the size of the weight matrices and the cache size of the CPU, we can store the matrices in sequential blocks using the Z-curve order \cite{zorder}. This approach improves spatial locality, and thus cache usage, minimizes false sharing when using multiple threads, and minimizes TLB misses. 


We utilize a model similar to \cite{Yotov:05} for optimizing cache performance by dividing the computation into Mini-GEMVs. Specifically, we partition the input and output vectors and the weight matrix into blocks of sizes $m_b \times 1$ and $m_b \times \bits t_b / 32$, respectively. The term $\bits t_b / 32$ accounts for packing multiple values into a single 32-bit integer. 
For temporal locality, we set the block dimensions such that 
\begin{equation} \label{eq:model0}
32 m_b + \bits m_b t_b + 32 t_b \leq \gamma, 
\end{equation} 
where $\gamma$ represents the size in bits of the L1 data cache. 


\mypar{Micro-GEMV} In the next step aim for an efficient innermost loop of the computation. In particular, we use different instructions based on the specification of the targeted CPU. The specification may include the available instructions (such as AVX2, AVX512, or AMX), the number of floating-point ports, and the number of available FMA (fused multiply-add) ports. By tailoring the instructions to the specific CPU, we can maximize register usage and instruction-level parallelism, leading to improved performance during computation.

\begin{figure}
    \centering
    \includegraphics[scale=0.9]{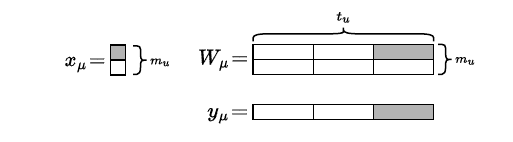}
    \caption{Visual representation of data used in the Micro-GEMV step. To store $x_\mu$, $W_\mu$, and $y_\mu$ we need $m_u$, $m_ut_u$ and $t_u$ registers, respectively. In each step of the algorithm we multiply one element of $x_\mu$ by a register containing $8$ values of $W_\mu$, to obtain $8$ values of $y_\mu$. Considering the grey cells, we have $y_{\mu,3} = y_{\mu,3} + x_{\mu,1} \cdot W_{\mu,1,17:24}$.}
    \label{fig:micro_gemv}
\end{figure}

In addition, we divide the Mini-GEMVs further into Micro-GEMVs operating on an $m_u$-sized part $x_\mu$ of $x$, $t_u$-sized part $y_\mu$ of $y$ and thus an $mu\times t_u$ block $W_\mu$ of $W$ as shown in Fig.~\ref{fig:micro_gemv}. $x_\mu$ and $W_\mu$ need to load different values, while $y_\mu$ remains in registers. We show a sketch of the modified innermost loop in~\cref{alg:muGEMV}. We set
\begin{equation} \label{eq:model1}
m_u + m_u t_u + t_u \leq \eta, 
\end{equation} 
where $\eta$ is the number of vector registers, to reduce the number of register spills.
For example, AVX/AVX2 have $\eta = 16$.

Finally, we include $m_u$ and $t_u$ in~\eqref{eq:model0} to simplify generation and reduce cleanup code. The final model equation is therefore
\begin{equation} \label{eq:model2}
\begin{aligned}
32 m_b + \bits m_b t_b + 32 t_b \leq \gamma, \\
t_b \mod t_u = 0,\quad m_b \mod m_u = 0.
\end{aligned}
\end{equation} 
We find $m_u$ and $t_u$ through empirical search. With these fixed, we maximize the left-hand side of the inequality to maximize cache utilization.

\begin{algorithm}
\small
\renewcommand\algorithmiccomment[1]{%
  \quad  // {#1} %
}
\caption{AVX2 4-bit $\mu$qGEMV Routine.}\label{alg:muGEMV}
\begin{algorithmic}[1]
\STATE $y_{1} \gets \avx{load}(\&y[j])$
\STATE{\dots}\COMMENT{Load the values of $y$ in $t_u$ registers, and use them as vector accumulators.}
\STATE $y_{t_u} \gets \avx{loadu}(\&y[j+t_u])$
\FOR{$i = 0:m_u:m$}
\STATE $w_{1,1} \gets \avx{load}(\&W[i/8][j])$
\STATE{\dots}\COMMENT{Load the values of $W$ in $t_um_u$ registers.}
\STATE $w_{t_u,m_u} \gets \avx{load}(\&W[(i+m_u)/8][j+t_u])$
\STATE $x_{1} \gets \avx{broadcast}(\&x[i])$
\STATE{\dots}\COMMENT{Load the values of $x$ in $m_u$ registers.}
\STATE $x_{m_u} \gets \avx{broadcast}(\&x[i+m_u])$
\STATE{\dots}\COMMENT{Compute}
\ENDFOR
\STATE $\avx{store}(\&y[j], y_{1})$
\STATE{\dots}\COMMENT{Store back $t_u$ registers.}
\STATE $\avx{store}(\&y[j+t_u], y_{t_u})$
\end{algorithmic}
\end{algorithm}

Finally, we present an overview of the code generation meta-algorithm below. 

\begin{algorithm}
\small
\caption{QIGen Generation Overview.}\label{alg:overview}
\begin{algorithmic}[1]
\STATE{Fetch $\gamma$ and $\eta$}
\STATE{Find $m_u$ and $t_u$ by optimizing~\eqref{eq:model1} via search.}
\STATE{Find $m_b$ and $t_b$ by maximising the left-hand side of~\eqref{eq:model2} using integer-programming.}
\STATE{Generate the Micro-GEMV kernels for the CPU.}
\end{algorithmic}
\end{algorithm}

\section{Evaluation}

We assess the effectiveness and precision of our implementation by comparing it with the Python bindings for llama.cpp~\cite{llamacpp}\footnote{https://github.com/abetlen/llama-cpp-python}, and by presenting the perplexity values on the standard \texttt{wikitext2} dataset~\cite{wikitext103}. For this preliminary version, we have executed our generator on the AVX2 instruction set. However, the instructions that we use have equivalents on all SIMD vector architectures. 

\mypar{Goals and setup}
We compare our approach to llama.cpp, both in terms of inference throughput (tokens generated / second) as well as in terms of accuracy of the resulting models, measured in terms of perplexity (PPL).  We use models generated using the GPTQ quantization method~\cite{frantar2022gptq}, whereas llama.cpp uses pre-generated models using their custom \texttt{q4\_0} quantization format. 
For performance measurements we use an AMD EPYC 7742 64-Core processor running $64$ threads. We compile our code using \texttt{gcc 9.4.0} with \texttt{-O3, -mavx, -mavx2, -mfma, -march=native, -ffast-math, -ftree-vectorize} flags, and we parallelize using \texttt{OpenMP 4.5}. We compile llama.cpp using OpenBLAS. 

\mypar{Accuracy} 
We begin by examining the accuracy (perplexity) of our generated models on the \texttt{wikitext2} dataset, which is standard in this setting~\cite{frantar2022gptq, yao2022zeroquant}. 
Moreover, Dettmers and Zettlemoyer have shown that perplexity is closely correlated with average performance across zero-shot tasks~\cite{dettmers2022case}. 
The results for running GPTQ with standard parameters across the LLaMA model family are shown in Table~\ref{table:ppl}. 

\begin{table}[h]
\centering
\scalebox{0.65}{
\begin{tabular}{@{}rrrrcrrrcrrr@{}}\toprule
& \multicolumn{3}{c}{llama-7b (5.68)} & \phantom{a}& \multicolumn{3}{c}{llama-13b (5.09)} &
\phantom{a} & \multicolumn{3}{c}{llama-30b (4.10)}\\
\cmidrule{2-4} \cmidrule{6-8} \cmidrule{10-12}
Group size & 4bit & 3bit & 2bit && 4bit & 3bit & 2bit && 4bit & 3bit & 2bit\\ \midrule
128 & 5.81 & 6.43 & 23.58 &&  5.19 & 5.51 & 15.88 && 4.21 & 4.63 & 11.13\\
64 &  5.79 & 6.23 & 14.15 &&  5.18 & 5.49 & 11.13 && 4.19 & 4.55 & 9.10\\
32 &  5.77 & 6.11 & 10.24 &&  5.15 & 5.40 & 8.37  && 4.18 & 4.47 & 7.22\\
16 &  5.76 & 5.99 & 8.30  &&  5.13 & 5.31 & 7.11  && 4.16 & 4.39 & 6.19\\
\bottomrule
\end{tabular}

}
\caption{Perplexity values evaluated on the \texttt{wikitext2} dataset. The perplexity of the floating point model is given in brackets next to their name.}
\label{table:ppl}
\end{table}

We observe that 4bit quantization generally preserves accuracy, across all models within a few relative percentage points, and that results improve when decreasing group size. 
(The only exception is LLaMA-7b group size 32, for which we believe some additional hyper-parameter tuning may be beneficial.) As also noted in the original paper, the relative accuracy drop decreases with model size, to the point where small group sizes are nearly lossless on the larger models. 

\begin{figure*}
    \centering
    \subfigure[]{
        \centering
        \includegraphics[scale=0.53]{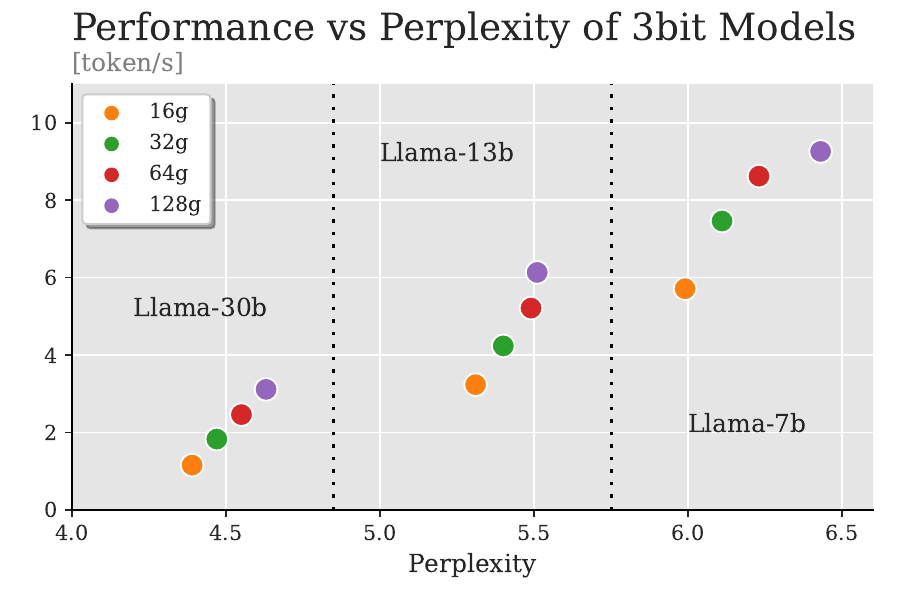}
        \label{subfig:perf_vs_ppl_3bit}
    }
\centering    
    \subfigure[]{
        \centering
        \includegraphics[scale=0.53]{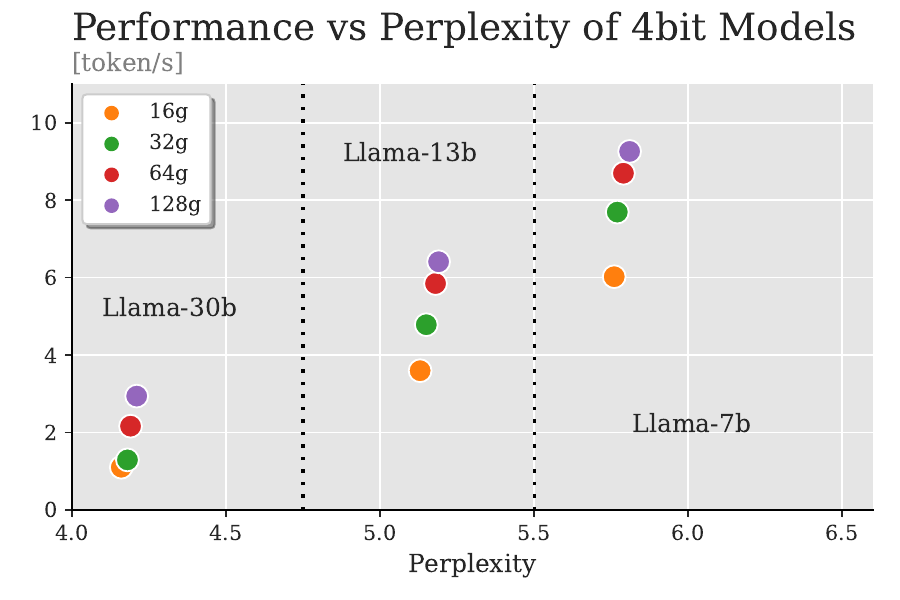}
        \label{subfig:perf_vs_ppl}
    }
    \caption{Changes in performance and perplexity between different group sizes used. In~\ref{subfig:perf_vs_ppl_3bit} and in~\ref{subfig:perf_vs_ppl} we give the measurements for our 3bit and 4bit implementations respectively. Overall we do not notice major performance difference between 3bit and 4bit kernels.}
    \label{fig:pvppl}
\end{figure*}

\begin{figure}[h!]
    \centering
    \includegraphics[scale=0.53]{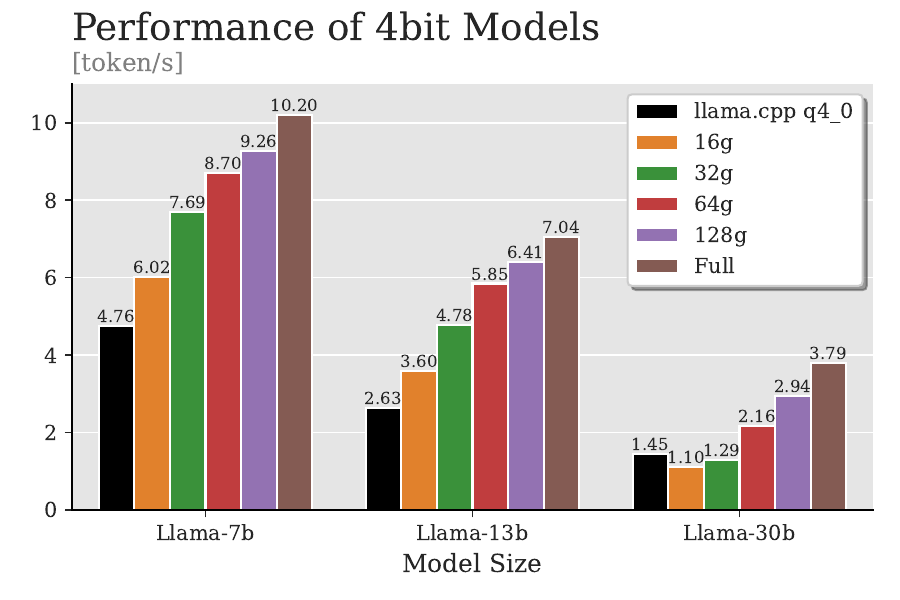}
    \caption{Performance comparison in $token/seconds$ of our different implementations, grouped and non grouped, vs.~llama.cpp $q4\_0$ quantization method. We notice performance benefits arising from bigger groups.}\label{fig:perf}
\end{figure}

\mypar{Throughput performance}
Next, we compare the performance of our generated kernels to those of llama.cpp. 
For a fair comparison, we consider models quantized to a 4bit format. 
For llama.cpp, we specifically executed their \texttt{q4\_0} quantization format, which yields the best performance among all their supported formats. (According to their documentation, this would correspond to a group size of 32 using our implementation.) 

We generate different variants for our kernels, varying the grouping between 32 and 128 elements, as well as using full columns, trading off accuracy for space. For full columns, we can use Equation~\eqref{eq:core} to reduce the number of floating point operations and loads in the innermost loop. On the other hand, for grouped models, we need to use Equation~\eqref{eq:core_group}. Note that, while Equation~\eqref{eq:core} requires storing the values of $y$ only once, Equation~\eqref{eq:core_group} requires $n/g$ additional stores, where $n$ is the number of rows and $g$ is the group size. 

The results are shown in Fig.~\ref{fig:perf}. We observe that our kernels outperform llama.cpp in terms of throughput, by up to $2.6 \times$ when using full columns on the $13$ billion parameter model, and by approximately $2 \times$ when using group implementations. However, we observed a lower performance than llama.cpp for the 30b case for group sizes 16 and 32. We believe that this is due to the assumptions we introduce to solve~\eqref{eq:model2}, and should be improvable by further examination. 

We investigate the full accuracy/performance trade-off in Fig.~\ref{fig:pvppl}, where we show the relationship between the performance in tokens/seconds and the perplexity of our 3bit and 4bit models in Fig.~\ref{subfig:perf_vs_ppl_3bit} and Fig.~\ref{subfig:perf_vs_ppl} respectively.  We found that for smaller group sizes, performance decreased while accuracy increased. It is important to note that overall, the performance between the two implementations is similar.

\begin{table}\centering
\ra{0.5}
\begin{tabular}{@{}rrrr@{}}\toprule
& Version & Group Size & Memory (MiB) \\
\midrule
\emph{LlaMA} $7b$
& FP32  &  -  & 26555\\
& 3 Bit & 128 & 8072\\
& 4 Bit & 128 & 8814\\ 
\midrule
\emph{LlaMA} $13b$
& FP32  &  -  & 50712\\
& 3 Bit & 128 & 14333\\
& 4 Bit & 128 & 16483\\ 
\midrule
\emph{LlaMA} $30b$
& FP32  &  -  & 125614\\
& 3 Bit & 128 & 31173\\
& 4 Bit & 128 & 37176\\ 
\bottomrule
\end{tabular}
\caption{Comparison of memory usage in MiB used to generate a $128$ token sentence.}\label{table:memory}
\end{table}

\mypar{Memory consumption}
Finally, we report in Table~\ref{table:memory} the total memory in MiB used to generate a $128$ token long sentence using floating point weights and our 3bit and 4bit quantized kernels with a group size of $128$. The results were obtained for LlaMA models with 7, 13, and 30 billion parameters. The 3bit kernels showed a reduction of up to $4$x, while the 4bit kernels showed a reduction of up to $3.3$x times compared to the floating point implementations.

\section{Discussion}

We provided evidence that an automatic code generation approach can yield strong results for quantized inference over large language models. 
Our results show that one can obtain state-of-the-art CPU inference performance using our methods, with minimal accuracy loss when compared to the uncompressed baseline. Our results can be extended along several directions: improving practical performance of existing kernels through fine-tuning, as well as targeting different CPU architectures and accelerator hardware. 





\bibliography{bibliography}
\bibliographystyle{icml2022}

\end{document}